\documentclass[letterpaper, 10 pt, conference]{ieeeconf}  

\IEEEoverridecommandlockouts                              

\usepackage{amsmath,amssymb,amsfonts}
\usepackage{graphicx} 
\usepackage[dvipsnames]{xcolor}
\usepackage{algorithm}
\usepackage{booktabs} 
\usepackage{algpseudocode}
\usepackage{bbold}
\usepackage{multirow}
\usepackage{svg}
\usepackage{subcaption}

\makeatletter
\let\NAT@parse\undefined
\makeatother
\usepackage{hyperref}
\hypersetup{
    colorlinks=true,
    linkcolor=blue,
    filecolor=magenta,      
    urlcolor=cyan,
    citecolor=red,
    pdftitle={goslam},
    pdfpagemode=FullScreen,
    }

\title{\LARGE \bf
Go-SLAM: Grounded Object Segmentation and Localization with Gaussian Splatting SLAM
}

\author{Phu Pham$^{1}$, Dipam Patel$^{1}$, Damon Conover$^{2}$,  Aniket Bera$^{1}$\\
$^1$Department of Computer Science, Purdue University  $^2$DEVCOM Army Research Laboratory\\
\texttt{\{phupham, dipam, aniketbera\}@purdue.edu, damon.m.conover.civ@army.mil}
}

\begin{document}

\maketitle
\thispagestyle{empty}
\pagestyle{empty}

\begin{abstract}
We introduce Go-SLAM, a novel framework that utilizes 3D Gaussian Splatting SLAM to reconstruct dynamic environments while embedding object-level information within the scene representations. This framework employs advanced object segmentation techniques, assigning a unique identifier to each Gaussian splat that corresponds to the object it represents. Consequently, our system facilitates open-vocabulary querying, allowing users to locate objects using natural language descriptions. Furthermore, the framework features an optimal path generation module that calculates efficient navigation paths for robots toward queried objects, considering obstacles and environmental uncertainties. Comprehensive evaluations in various scene settings demonstrate the effectiveness of our approach in delivering high-fidelity scene reconstructions, precise object segmentation, flexible object querying, and efficient robot path planning. This work represents an additional step forward in bridging the gap between 3D scene reconstruction, semantic object understanding, and real-time environment interactions.

\end{abstract}

\section{INTRODUCTION}

Autonomous robots are becoming increasingly vital in various fields, including search and rescue, manufacturing, and military operations \cite{Queralta-SAR, Drew-SAR, Krnjaic-warehouse, Cheraghi-swam}. To effectively navigate and interact with their environment, these robots need the ability to accurately reconstruct the surroundings, segment objects of interest, and plan paths in real-time. One of the key challenges in building such systems lies in achieving high-fidelity scene reconstruction while also integrating semantic understanding of objects within the scene. Additionally, enabling robots to query objects in an open-vocabulary manner and generate optimal paths to interact with these objects enhances their flexibility and adaptability in challenging environments.

Traditional SLAM or Simultaneous Localization and Mapping techniques \cite{Cadena-slam, Campos-slam, Droid-slam} have proven effective in reconstructing environments but often fail to provide detailed, object-level segmentation and interaction capabilities. In contrast, methods like point cloud or voxel-based reconstructions, while offering spatial accuracy, tend to struggle with incorporating object semantics in a robust and scalable manner. Recent advances in 3D Gaussian Splatting \cite{3dgs} offer a promising alternative for scene representation and rendering by using 3D Gaussian primitives to model the geometry and appearance of a scene.

While accurate 3D reconstruction is essential, true scene understanding requires the ability to identify and label objects within the environment. To address this, we incorporate advanced computer vision models that provide robust object detection and precise segmentation capabilities. By leveraging these techniques with 3D Gaussian Splatting, we generate a semantically rich environmental representation, where each Gaussian splat is associated with an object label. This enables robotic systems to understand both the spatial structure of the environment and the semantic relationships between objects, allowing for accurate object identification, tracking, and interaction across multiple camera frames.

\begin{figure}
    \centering
    \includegraphics[width=\linewidth]{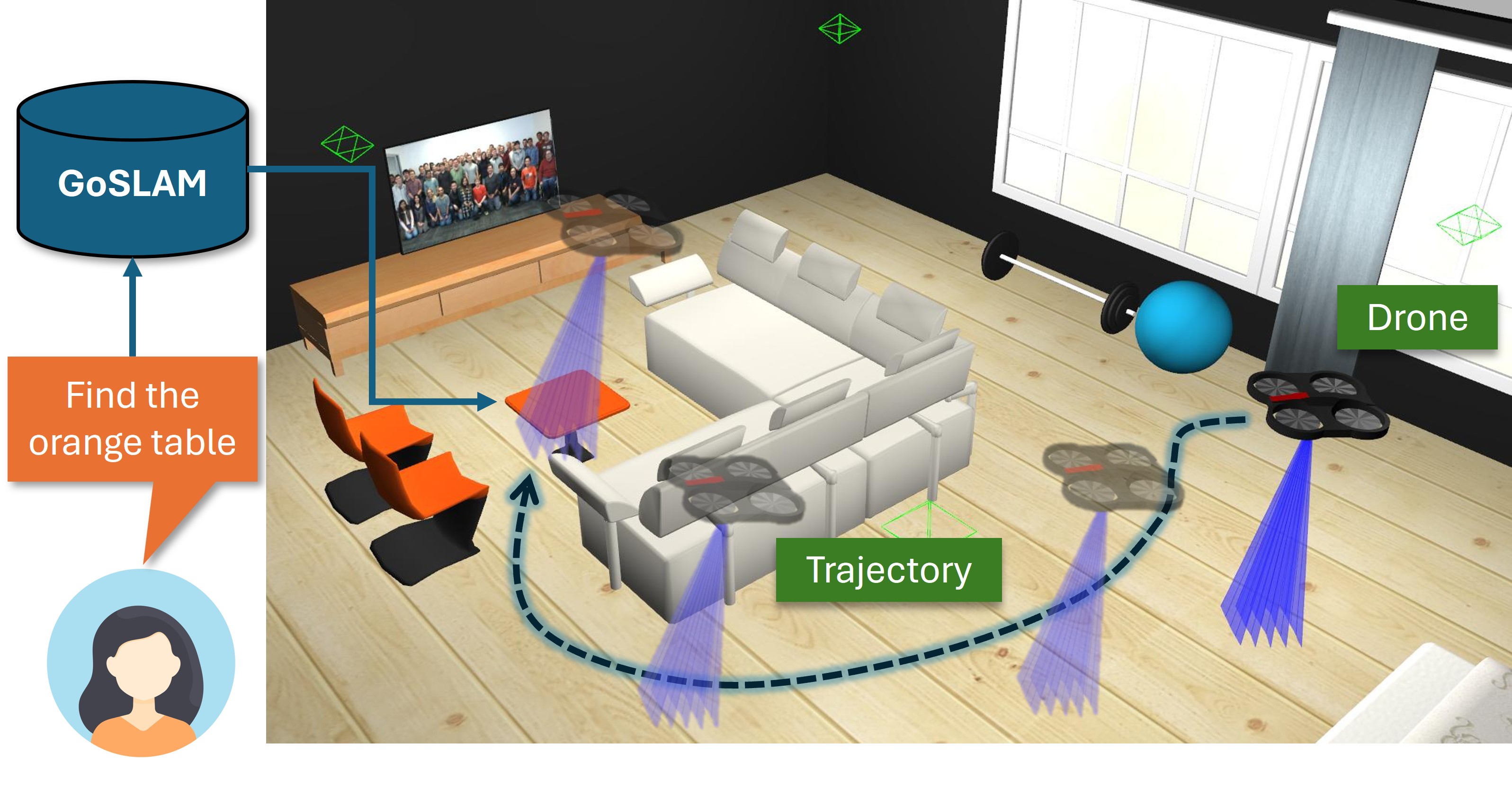}
    \caption{The entire pipeline of Go-SLAM -- The user queries the 3D reconstructed model in real-time with a specific object in the environment. The Go-SLAM detects the queried object and provides the 3D world coordinates of the goal location. The drone now navigates to the provided co-ordinates using PRM path planning algorithm}
    \label{fig:teaser}
\end{figure}

Another novelty of our approach is the support for open-vocabulary queries. By incorporating natural language processing techniques, our system allows users or higher-level planning algorithms to locate objects using flexible, human-like descriptions. This capability significantly enhances the adaptability of robotic systems, enabling them to understand and act upon a wide range of commands without being limited to a predefined set of object categories. Our entire approach is outlined in Figure \ref{fig:teaser}.

Finally, we demonstrate the practical utility of our framework by implementing an optimal path-planning algorithm that leverages the semantically annotated 3D model. This allows a robot to efficiently navigate from its current position to a queried object, taking into account the spatial layout and potential obstacles in the environment.

The main contributions of this paper can be summarized as follows:
\begin{itemize}
    \item A novel implementation of 3D Gaussian Splatting SLAM with state-of the-art object segmentation and labeling techniques.
    \item An open-vocabulary querying system that enables flexible object localization in 3D reconstructed environments.
    \item Comprehensive experimental results demonstrate the effectiveness of our approach, showing improvements in precision, recall, and IoU by up to 17\%, 27\%, and 35\%, respectively, across various scenarios.
\end{itemize}

By combining these components, our framework represents a significant step towards creating more intelligent and adaptable robotic systems. To the best of our knowledge, this is the first SLAM framework capable of both understanding and interacting with complex, unknown environments.

\section{RELATED WORK}

\subsection{3D Gaussian Splatting}

3D Gaussian Splatting (3DGS) has emerged as a powerful technique for representing and rendering 3D scenes. Originally introduced by Kerbl et al.\cite{3dgs}, 3DGS uses a collection of 3D Gaussian primitives to model geometry and appearance. This approach offers several advantages over traditional 3D reconstruction methods. Unlike mesh-based  \cite{PoissonRecon, MarchingCube}, which can be computationally expensive and complex to manipulate, Gaussian Splatting provides a more flexible representation by modeling surfaces with continuous, parameterized Gaussians. This allows for smoother surface approximations, especially for objects with irregular or complex geometries. Additionally, Gaussian Splatting can handle partial or noisy data more robustly, making it well-suited for real-time applications such as SLAM, where incomplete or uncertain information is common. The method also facilitates efficient integration of multimodal data, such as combining RGB and depth information, which enhances the accuracy of scene reconstructions. 

\subsection{SLAM systems}

Simultaneous Localization and Mapping (SLAM) is a fundamental problem in robotics and computer vision, aiming to construct a map of an unknown environment while simultaneously tracking an agent's location within it. Traditional SLAM approaches have relied on sparse feature-based methods \cite{Rublee-orb} or dense volumetric representations \cite{kineticfusion}. More recently, neural implicit representations like Neural Radiance Fields (NeRF) \cite{NERF} have been adapted for tasks like SLAM \cite{nerf-slam} and navigation \cite{10342420}, offering high-quality scene reconstruction, but often at the cost of computational efficiency \cite{Point-SLAM, NICE-slam, nerf-slam, iMAP}.

\subsection{Gaussian Splatting SLAM}

The integration of 3D Gaussian Splatting into SLAM systems is a recent development that aims to leverage the advantages of 3DGS for real-time mapping and localization. Matsuki et al. \cite{MonoGS} introduced the first Gaussian Splatting SLAM system, demonstrating its effectiveness in monocular settings. GS-SLAM by Yan et al. \cite{gs-slam} proposed a dense visual SLAM system using 3DGS, achieving competitive performance in both reconstruction and localization with lower time consumption compared to other methods. SplaTAM by Keetha et al. \cite{splatam} introduced a system for dense RGBD SLAM using 3DGS, demonstrating real-time performance and high-quality reconstruction.

These Gaussian Splatting SLAM approaches have shown promising results in terms of reconstruction quality, localization accuracy, and computational efficiency. However, challenges remain in areas such as large-scale mapping, loop closure, and handling dynamic environments.

\subsection{Object detection and segmentation}

Object detection and segmentation are essential for enabling robots to understand and interact with their environment. Traditional approaches, such as Faster R-CNN \cite{frcnn} and Mask R-CNN \cite{mask-rcnn} have been instrumental in detecting objects and generating instance-specific segmentation masks, enabling robots to recognize and differentiate objects within complex scenes. In addition, methods such as YOLO (You Only Look Once) \cite{yolo} introduced single-shot detectors that achieve real-time object detection by predicting bounding boxes and class probabilities directly from full images, significantly improving the efficiency of object recognition tasks. 

More recent models, such as Grounding DINO \cite{groundingdino} and Segment Anything Model (SAM) \cite{sam}, have pushed the boundaries of object segmentation. Grounding DINO uses transformer-based architectures to detect and localize objects by understanding semantic context, while SAM excels at universal segmentation by generating masks for any object given minimal input, without needing retraining. Grounded SAM, which integrates Grounding DINO with SAM, enables detection and segmentation of any objects specified by an input text prompt. These models empower robotic systems to recognize and segment objects flexibly, even in environments with undefined or unseen objects.

\subsection{Language embedded for 3D reconstruction}

Recent developments in 3D scene reconstruction have increasingly incorporated language understanding, enabling more intuitive interactions with 3D environments. A notable work in this field is LERF (Language Embedded Radiance Fields) \cite{lerf2023}, which embeds CLIP-based language representations into Neural Radiance Fields (NeRF). LERF constructs a dense, multi-scale language field by rendering CLIP embeddings along training rays, facilitating zero-shot, pixel-aligned queries without the need for region proposals or masks. This approach supports real-time generation of 3D relevancy maps for diverse language prompts, offering potential applications in robotics, vision-language model analysis, and interactive scene exploration.

\begin{figure*}[htbp]
    \centering
    \includegraphics[width=\linewidth]{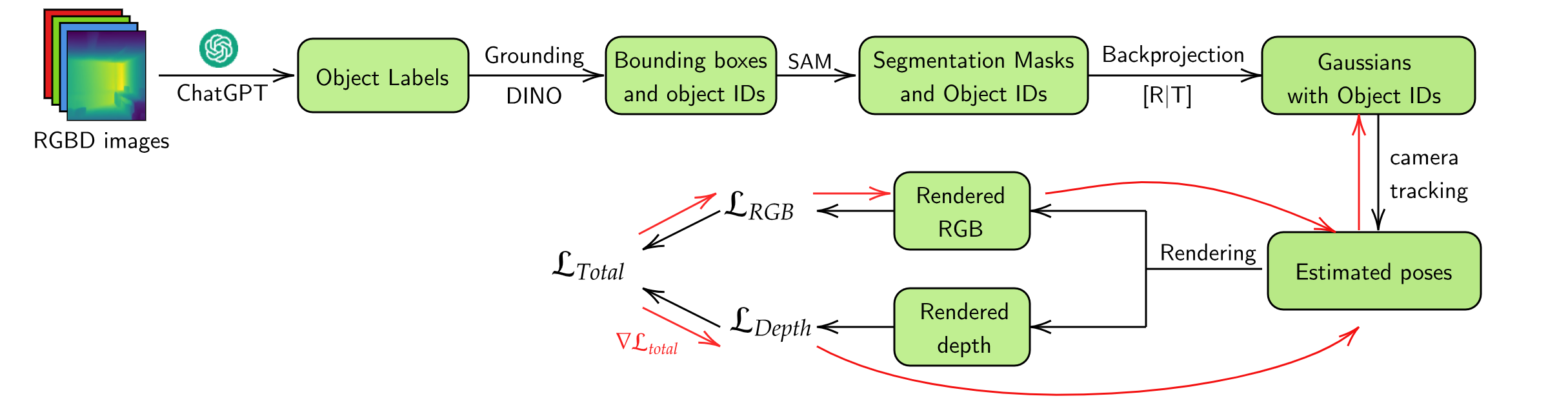}
    \caption{Overview of our Go-SLAM framework for environment reconstruction and language embedded feature.}
    \label{fig:overview}
\end{figure*}

Building upon LERF, LangSplat \cite{Langsplat} offers a more efficient approach for creating 3D language fields using 3D Gaussian Splatting rather than NeRF. By encoding CLIP-based language features \cite{clip} into 3D Gaussians, LangSplat significantly reduces computational cost through a tile-based splatting method. Additionally, it incorporates a language autoencoder to lower memory usage and uses SAM for hierarchical semantics learning, improving object boundary precision. These advancements lead to faster and more accurate open-vocabulary 3D object localization and semantic segmentation compared to LERF.

\section{METHOD}\label{sec:method}

In this section, we outline the comprehensive methodology of our framework, namely Go-SLAM, which employs cutting-edge techniques to achieve high-precision and efficient 3D reconstruction of environments captured through RGBD cameras. To the best of our knowledge, this is the pioneering SLAM system to integrate language features, enabling open-vocabulary object detection and localization. Figure. \ref{fig:overview} illustrates the overview of our framework. In the following sections, we will elaborate on each component in the pipeline.

\subsection{3D Gaussian Splatting SLAM framework}

\begin{figure*}[htbp]
    \centering
    \begin{subfigure}{0.32\textwidth}
        \centering
        \includegraphics[width=\linewidth]{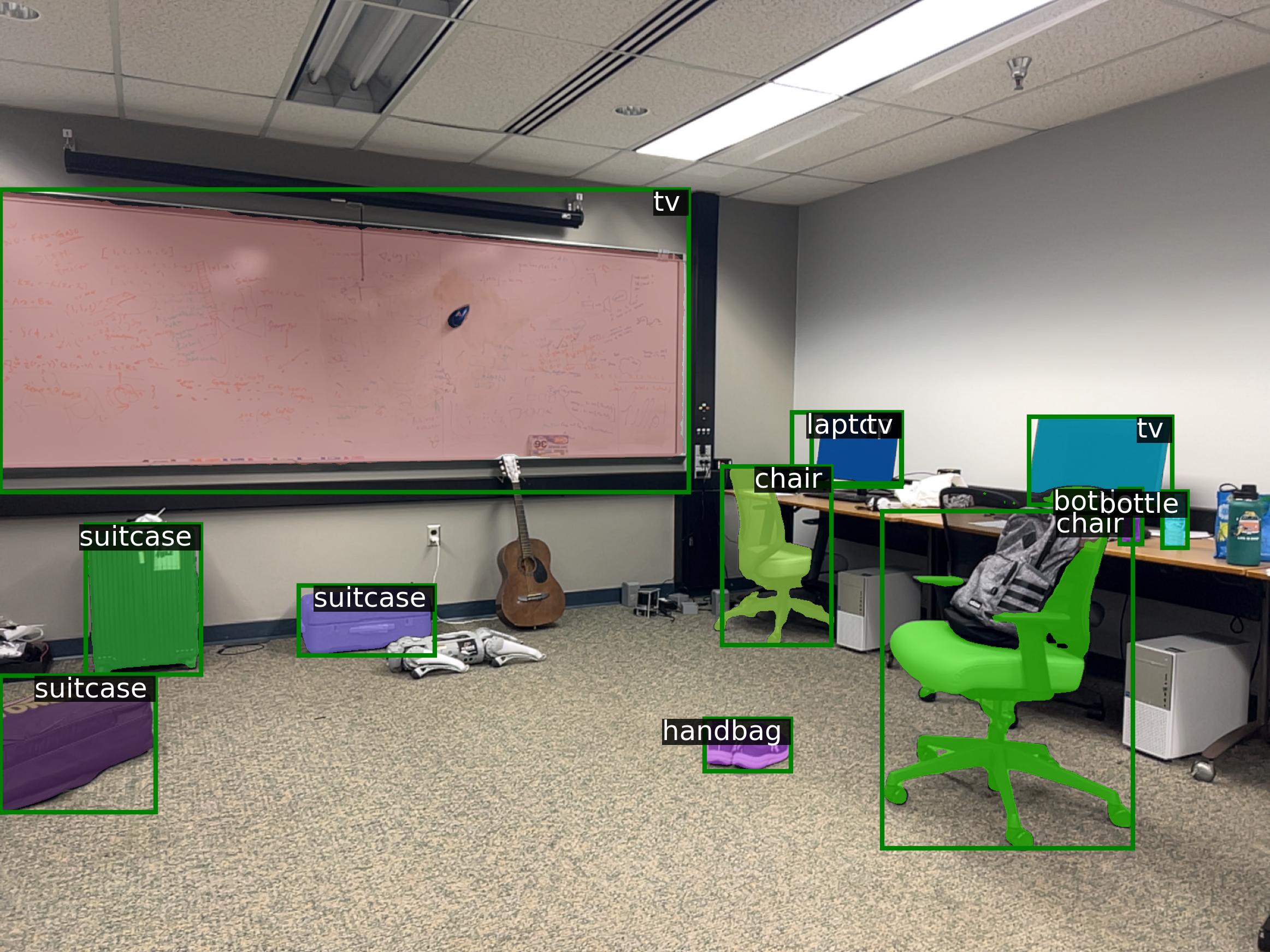}
        \caption{YOLO + SAM}
        \label{fig:image1}
    \end{subfigure}
    \hfill
    \begin{subfigure}{0.32\textwidth}
        \centering
        \includegraphics[width=\linewidth]{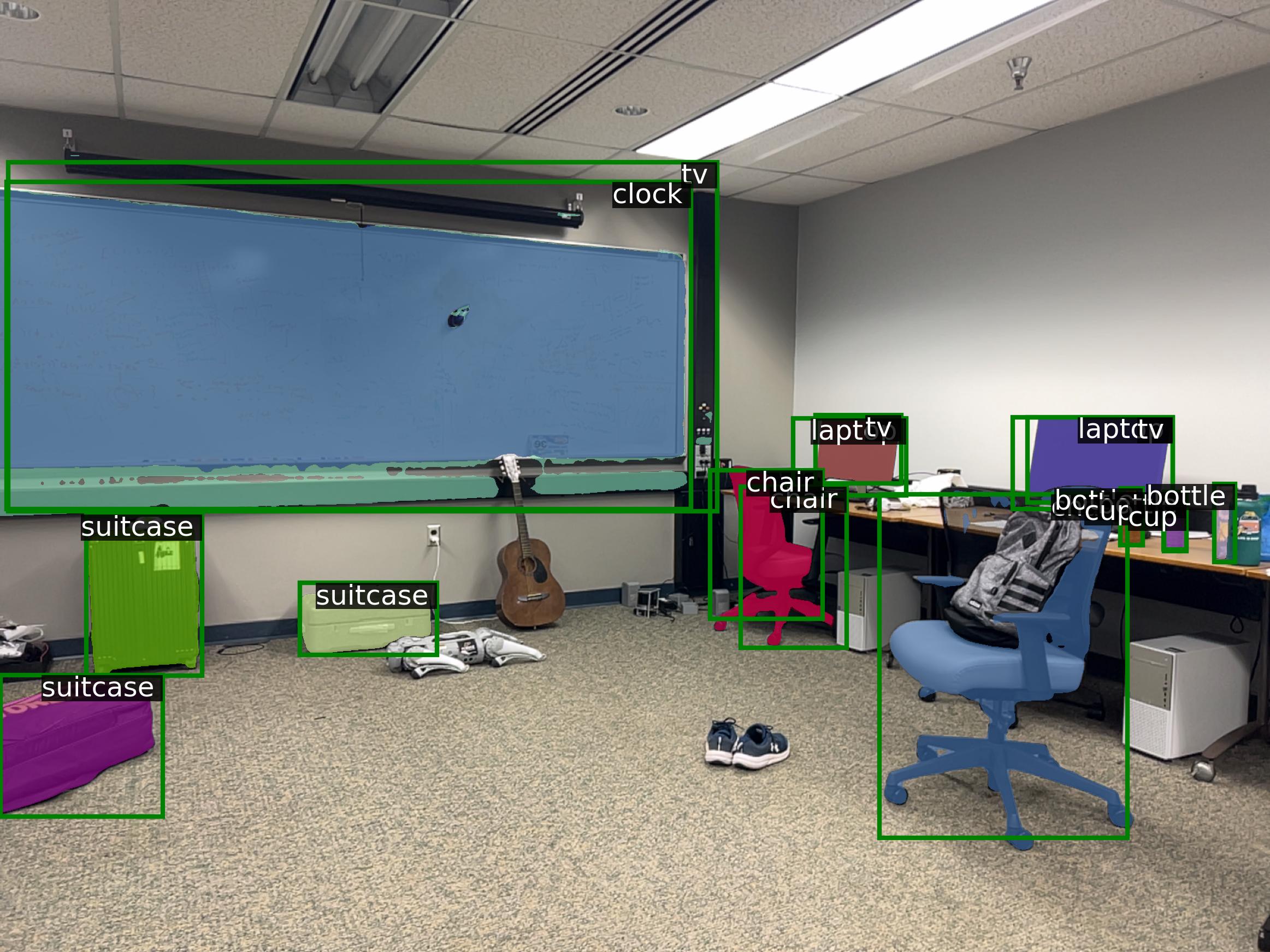}
        \caption{Faster RCNN + SAM}
        \label{fig:image2}
    \end{subfigure}
    \hfill
    \begin{subfigure}{0.32\textwidth}
        \centering
        \includegraphics[width=\linewidth]{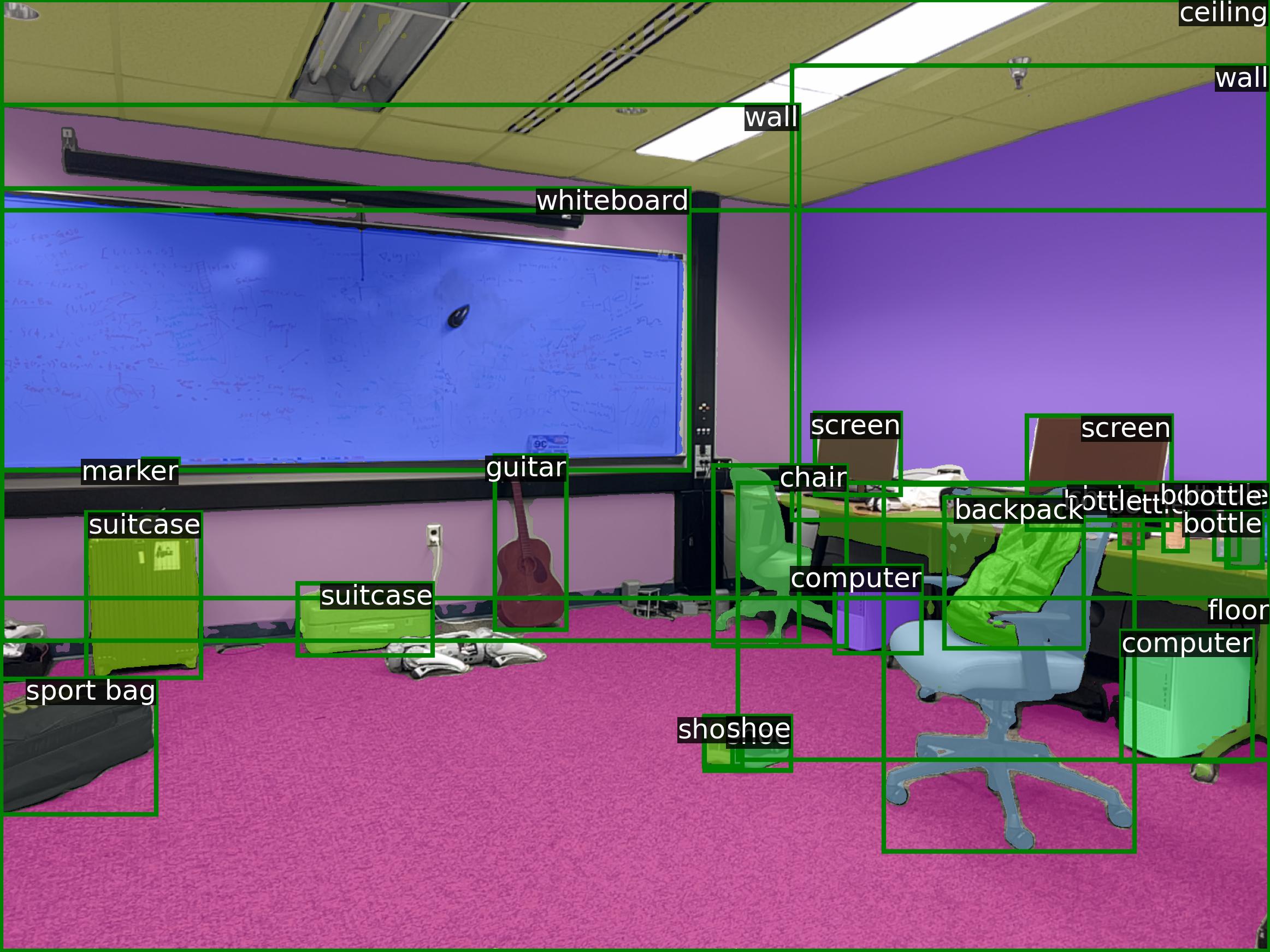}
        \caption{Grounding DINO + SAM}
        \label{fig:image3}
    \end{subfigure}
    \caption{Performance comparison between different object detectors for grounded object segmentation.}
    \label{fig:detector-comparison}
\end{figure*}

We employ 3D Gaussian Splatting to represent the reconstructed environment. Our proposed method builds upon the SplaTAM \cite{splatam} approach to implement a robust and efficient 3DGS SLAM system. We leverage the strengths of explicit volumetric representations using 3D Gaussians to enable high-fidelity reconstruction from a single RGBD camera.

\subsubsection{Differential 3D Gaussian Splatting representation}

The core concept behind 3D Gaussian splatting is to represent the scene as a collection of Gaussians, where each Gaussian splat encodes key attributes like position, scale, orientation, color, opacity, and object association. These Gaussians act as probabilistic volumetric representations of the scene, allowing us to efficiently approximate the underlying geometry and appearance from multiple viewpoints.

Each splat is modeled as a 3D Gaussian distribution with parameters:

\begin{equation}
    G(x) = c \cdot \exp \left( -\frac{1}{2} (x - \mu)^\top \Sigma^{-1} (x - \mu) \right)
\end{equation}

where $x \in \mathbb{R}^3$ is a 3D point in space, $\mu \in \mathbb{R}^3$ is the Gaussian center, representing the position of the splat, $\Sigma \in \mathbb{R}^{3 \times 3}$ is the covariance matrix, defining the scale and orientation of the splat in 3D space, and $c \in \mathbb{R}^3$ is the color vector (RGB) associated with the splat,

The exponential term models the spatial influence of the splat, decaying with distance from the center \(\mu\). Our framework uses this probabilistic approach to model both the geometry and appearance of the environment, with splats efficiently representing 3D surfaces across multiple viewpoints.

A key advantage of this 3DGS representation is that it allows for differentiable rendering, enabling us to optimize the parameters of each Gaussian (such as $\mu$, $\Sigma$, and $c$) through backpropagation. By rendering the scene from multiple viewpoints and comparing the rendered images with ground truth, we can compute a loss function, such as $L1$ or $L2$ loss, and backpropagate the error to adjust the parameters. This optimization process refines the Gaussian representations to better approximate the geometry and appearance of the scene, leveraging the differentiability of the rendering process to iteratively improve the 3D reconstruction.

\subsubsection{Tracking and Gaussian densification}

For each captured RGBD image, the framework back-projects each pixel $(u, v)$ with depth $d$ into 3D space using the camera's intrinsic matrix $K$, and converts it into a 3D Gaussian splat. The back-projection is computed as:

\begin{equation}
    X_c = d \cdot K^{-1} \begin{bmatrix} u \\ v \\ 1 \end{bmatrix}, \quad X_w = R_c X_c + t_c    
\end{equation}

where $X_c$ and $X_w$ are the point coordinates in camera and world frames, respectively, $R_c$ and $t_c$ are the rotation and translation matrices representing the camera's pose.

Each back-projected point is then converted into a Gaussian splat with its center $\mu$ set to $X_w$, and its color $c$ taken from the corresponding pixel in the RGB image. We employ the camera tracking method introduced by \cite{splatam} to estimate the camera pose for the current RGBD image. Camera parameters are optimized using gradient descent based on the $L1$ losses of the rendered colors and depths. Additionally, a silhouette mask is rendered to capture the density of the scene, which facilitates in quickly identifying previously mapped areas. This facilitates more efficient Gaussian densification for incoming RGBD images.

\subsection{Grounded object segmentation}

In our framework, object segmentation plays a critical role in embedding semantic information into the reconstructed 3D environment. We use Grounding DINO \cite{groundingdino} for object detection and SAM \cite{sam} for instance segmentation, as inspired by Grounded SAM \cite{groundedSAM}. Grounding DINO combines visual and language features to detect objects by grounding text-based queries in specific image regions, requiring input labels as text prompts to identify objects. SAM (Segment Anything Model), on the other hand, generates high-quality instance segmentations without predicting object labels. Instead, it focuses purely on producing object masks, which can then be assigned labels through external means. This reliance on text-based inputs for Grounding DINO becomes problematic in unknown environments where predefined labels may not be available, limiting the system’s autonomous capabilities.

Several existing models, such as Faster R-CNN \cite{frcnn}, YOLO \cite{yolo}, and DETR \cite{detr}, can directly provide object labels with bounding boxes. These models are highly effective in detecting objects and assigning labels based on the datasets they were trained on. However, a major drawback of these methods is their reliance on predefined class labels. They are typically trained on large datasets, such as COCO \cite{COCO} or ImageNet \cite{ImageNet}, which contain a fixed set of object categories. As a result, these models perform poorly in open and unknown environments, where new or unseen object classes may appear.


To overcome these limitations, we integrate ChatGPT 4o model to autonomously analyzes captured images and generates a list of object labels without requiring predefined prompts or fixed object categories. By utilizing its strong visual understanding capability, the model identifies a wide range of objects, providing corresponding labels based on the visual content of the image. This allows for open-vocabulary detection, even in previously unseen environments.

Formally, given an image $ I_i $, the ChatGPT 4o model processes the image and outputs a list of object labels $ \{L_j\} $, where $ L_j $ corresponds to an object $ O_j $ detected in the image. These labels are then passed to Grounding DINO, which assigns bounding boxes $ B_j $ and classifies objects based on the provided labels. SAM is subsequently used to generate pixel-level segmentation masks $ M_j $.

Given the segmentation mask, we assign unique object IDs to the corresponding 3D Gaussians in the scene. For each segmented object $ O_j $, the pixels within the mask $ M_j $ are back-projected into 3D space, and the corresponding 3D Gaussian splats are updated with the object ID $ C_j $.

For each object, $ O_j $ segmented in image $ I_i $, the object ID $ C_j $ is propagated to all Gaussians $ G_k $ corresponding to the pixels within the object's segmentation mask:

\begin{equation}
    C_j = \{G_k \mid (u_k, v_k) \in M_j\}
\end{equation}
This ensures that the semantic information from object segmentation is embedded directly into the 3D reconstruction, enabling effective querying and interaction with the scene.

To evaluate the performance of our proposed grounded object segmentation method, we conducted comparative experiments with two established object detectors, Faster R-CNN \cite{frcnn} and YOLO \cite{yolo}. The results, illustrated in Figure \ref{fig:detector-comparison}, indicate that our method is capable of detecting nearly all objects in the scene with accurate segmentation. Conversely, both YOLO and Faster R-CNN struggle to recognize objects outside their pretrained categories, such as `guitar', `backpack', and `computer'. Moreover, these models misclassify `whiteboard' and `desktop monitor' as `tv', `water bottle' as 'cup', and `sport bag' as `suitcase'. These findings demonstrate the robustness of our method in handling diverse and unknown environments.

\subsection{Open-vocabulary object queries}

\begin{figure*}
    \centering
    \includegraphics[width=\linewidth]{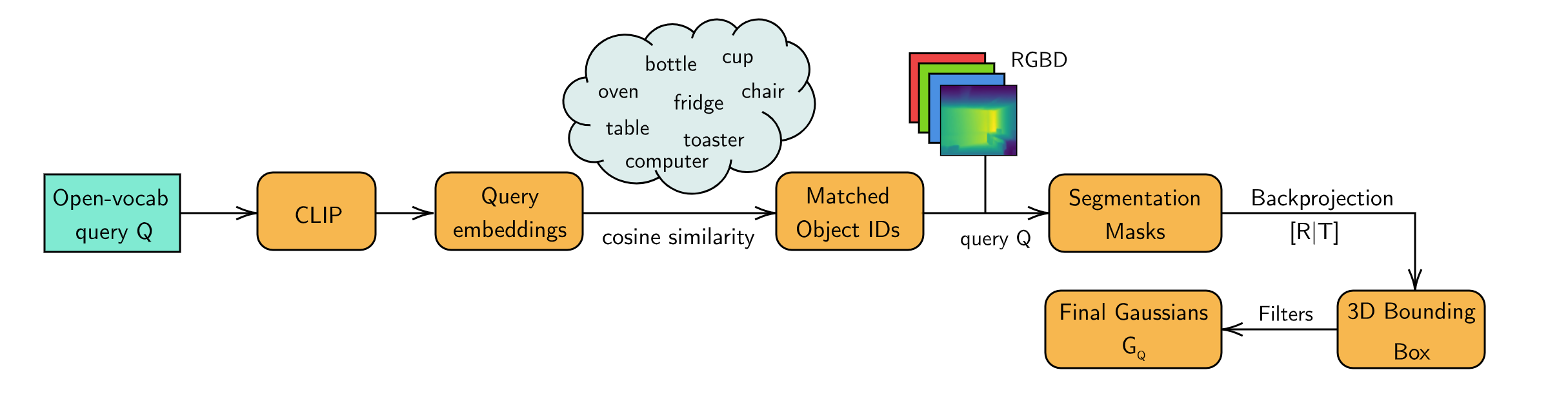}
    \caption{Open-vocabulary query pipeline}
    \label{fig:query}
\end{figure*}

Our framework supports open-vocabulary object querying, enabling users to search for objects in the 3D reconstructed environment using textual descriptions. The process involves matching the input query with detected object classes and refining the search to ensure precise identification of the queried object. The overall pipeline is shown in Figure. \ref{fig:query}.

\subsubsection{Query matching}

Given an input text query $Q$, we first compute the similarity between the CLIP embedding \cite{clip} of the input text and the CLIP embeddings of the detected object classes from the reconstructed environment. This allows us to determine the object class that most closely matches the input query based on semantic similarity.

For each detected object class \(C_j\), we compute the cosine similarity between the CLIP embedding of the input query \(E_Q\) and the embedding of the object class \(E_{C_j}\):

\begin{equation}
    s = \cos(E_Q, E_{C_j})
\end{equation}

The object class with the highest matching score $s$ is selected as the best match for the query.

\subsubsection{Pruning the search space}

Once the most similar object class is identified, we prune the search space to include only the keyframes that contain instances of this detected object class. This reduces the search complexity by narrowing the candidate frames where the queried object may be located. However, the detected object classes are typically in their most general form (e.g., "table" instead of "dining table" or "chair" instead of "swivel chair"). To ensure precise object identification, we rerun the grounded segmentation on these keyframes, refining the object boundaries to better match the specific object characteristics from the query. If the queried object is not found in the selected keyframes, the system expands the search to the remaining keyframes that were initially excluded. This ensures that objects potentially missed in the initial pruning phase are still considered during the search process.


\subsubsection{Object Localization}

Once the object matching the query is identified, we backproject all its segmented pixels into 3D space and compute a 3D bounding box that encompasses the object. This bounding box provides an approximate spatial boundary of the object within the reconstructed environment.

Next, we select all the 3D Gaussian splats that lie within this bounding box. Each Gaussian splat has an associated object ID, and the set of object IDs corresponding to the Gaussians inside the bounding box is denoted as $C_{\text{in}}$. Since the 3D bounding box may not fully capture the entire object, we expand the selection by including all Gaussians whose object IDs belong to $C_{\text{in}}$, ensuring that the complete object is included in the final set of Gaussians corresponding to the query $Q$:

\begin{equation}
    G_Q = \{G_k \mid \text{ID}(G_k) \in C_{\text{in}}\}
\end{equation}


For system evaluation and deployment, we utilized Gazebo (version 11.0) \cite{gazebo} in conjunction with ROS2 (Robot Operating System 2) \cite{doi:10.1126/scirobotics.abm6074} to simulate complex 3D environments. Our navigation strategy employs the PRM (Probabilistic Road Map) \cite{508439} path planning algorithm. This algorithm operates on a point cloud generated from the Gaussian centers derived from the reconstructed environment. The center of the object's bounding box, as determined by our object detection algorithm, serves as the goal point for path planning. This approach enables efficient obstacle avoidance and precise navigation within the reconstructed 3D space. By integrating these components, our system demonstrates robust performance in localizing queried objects and generating optimal paths for robot navigation, effectively bridging the gap between 3D scene reconstruction, semantic understanding, and real-time robotic interaction.

\begin{table*}[h]
\centering
\begin{tabular}{|c|c|c|c|c|c|c|c|c|c|}
\hline
\multirow{2}{*}{\textbf{Methods}} & \multicolumn{3}{c|}{\textbf{Office 2}} & \multicolumn{3}{c|}{\textbf{Room 0}} & \multicolumn{3}{c|}{\textbf{Room 2}} \\ \cline{2-10} 
                                  & \textbf{Precision} & \textbf{Recall} & \textbf{IoU} & \textbf{Precision} & \textbf{Recall} & \textbf{IoU} & \textbf{Precision} & \textbf{Recall} & \textbf{IoU} \\ \hline
Baseline 1                        & 0.38               & 0.42            & 0.33         & 0.40               & 0.39            & 0.31         & 0.39               & 0.41            & 0.32         \\ \hline
Baseline 2                        & 0.54               & 0.58            & 0.44         & 0.52               & 0.56            & 0.42         & 0.53               & 0.57            & 0.43         \\ \hline
Go-SLAM (ours)                    & \textbf{0.61}      & \textbf{0.73}   & \textbf{0.59} & \textbf{0.60}      & \textbf{0.70}   & \textbf{0.56} & \textbf{0.62}      & \textbf{0.72}   & \textbf{0.58} \\ \hline
\end{tabular}
\caption{Performance comparison of different methods across multiple environments.}
\label{tab:quantitative}
\end{table*}

\section{Experiments and results}

This section details the experimental setup, evaluation metrics, and results obtained from testing our Go-SLAM framework on different scene settings. The experiments are structured to evaluate the robustness and accuracy of our system in varied environments.

\subsection{Experimental Setup}

The experimental setup for assessing the effectiveness of our grounded object segmentation method included a variety of environments. For controlled testing, we utilized a subset of 18 scenes from the Replica dataset \cite{replica19}.

To establish a benchmark for our system, we conducted comparative evaluations against two baseline models. The first baseline utilized the Faster R-CNN segmentation model while maintaining the rest of our proposed pipeline. The second baseline incorporated our advanced grounded object segmentation coupled with 3D Gaussian Splatting SLAM (3DGS SLAM); however, it diverged from our full methodology by directly comparing the query against detected object labels, thereby omitting our carefully designed object matching algorithm. This approach allowed us to isolate the impact of our matching algorithm on the system's overall performance.


\subsection{Evaluation Metrics}

The evaluation of model performance utilized precision and recall metrics, which assess the accuracy of identifying relevant objects and the capability to detect all human-labeled ground truth 3D bounding boxes, respectively. Additionally, the Intersection over Union (IoU) was measured, calculating the overlap between the predicted bounding boxes and these human-labeled ground truth bounding boxes, thereby providing a quantitative measure of localization accuracy.

\subsection{Results}

The results of our experiments are evaluated both qualitatively and quantitatively to assess the effectiveness of our proposed grounded object segmentation method.

\subsubsection{Qualitative results}

\begin{figure}[ht!]
    \centering
    \begin{subfigure}[b]{0.49\columnwidth}
        \centering
        \includegraphics[width=\textwidth]{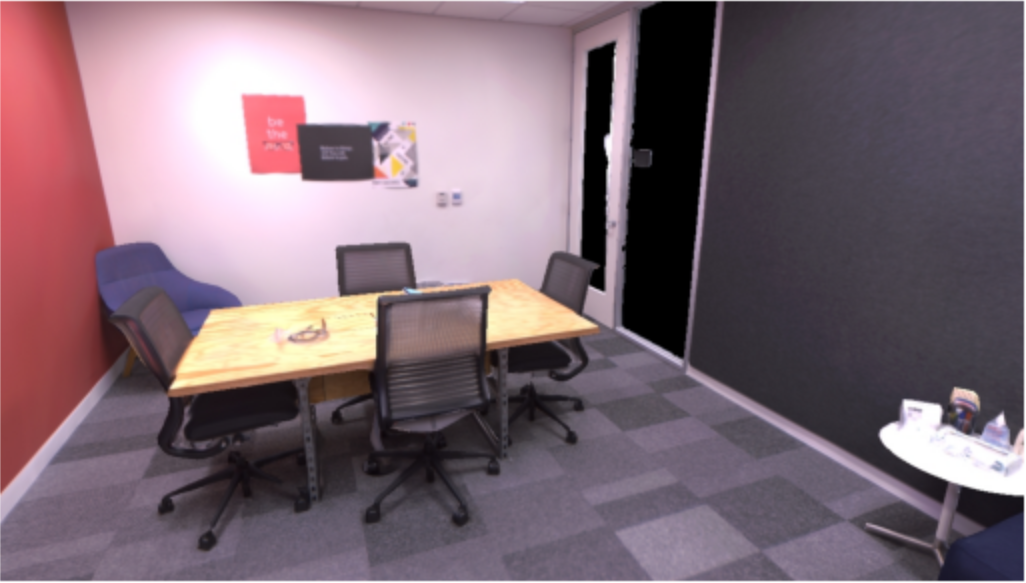}
        \caption{Ground truth RGB}
        \label{fig:figure1}
    \end{subfigure}
    \hfill
    \begin{subfigure}[b]{0.49\columnwidth}
        \centering
        \includegraphics[width=\textwidth]{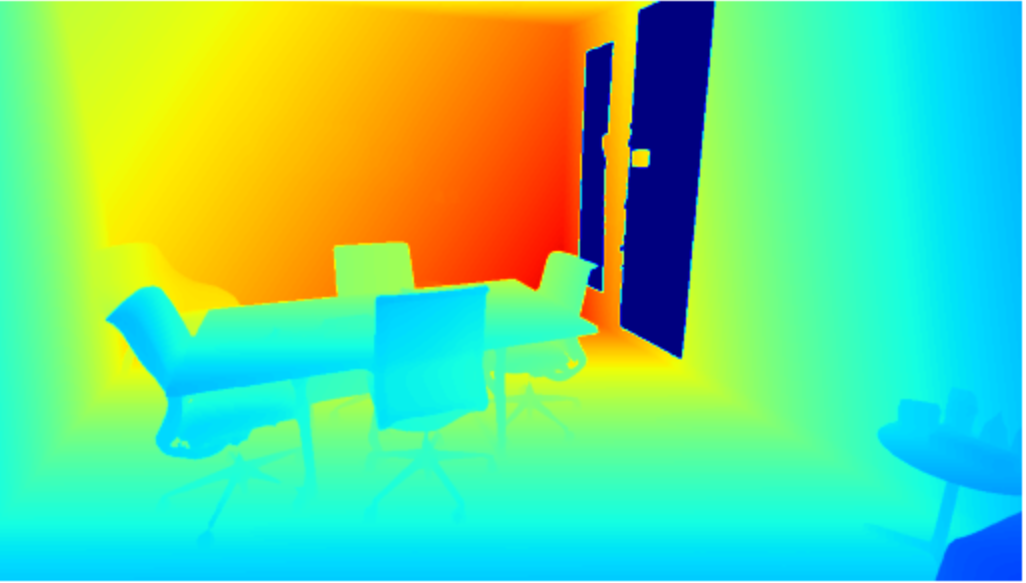}
        \caption{Ground truth depth}
        \label{fig:figure2}
    \end{subfigure}
    \vskip\baselineskip
    \begin{subfigure}[b]{0.49\columnwidth}
        \centering
        \includegraphics[width=\textwidth]{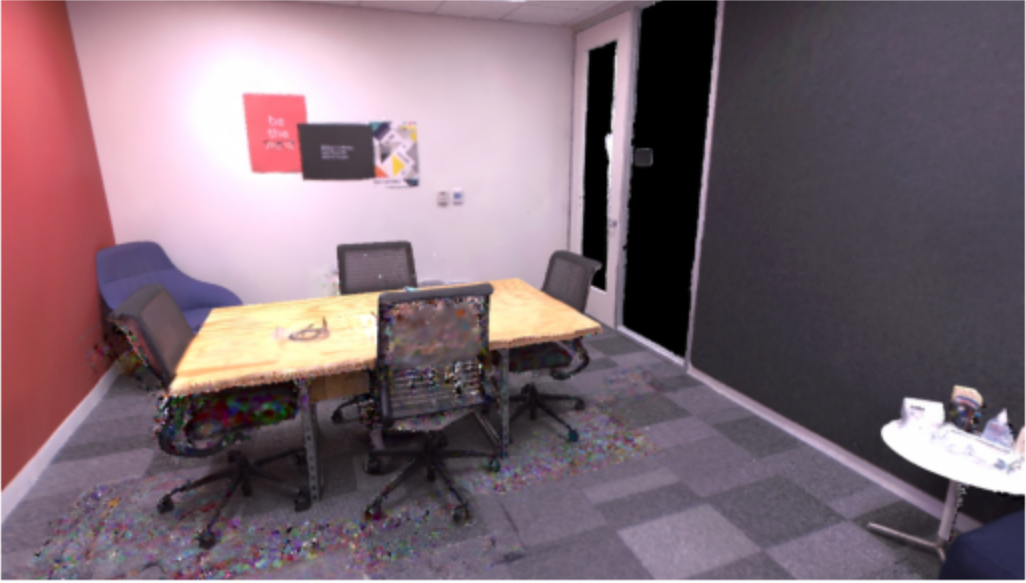}
        \caption{Rendered RGB}
        \label{fig:figure3}
    \end{subfigure}
    \hfill
    \begin{subfigure}[b]{0.49\columnwidth}
        \centering
        \includegraphics[width=\textwidth]{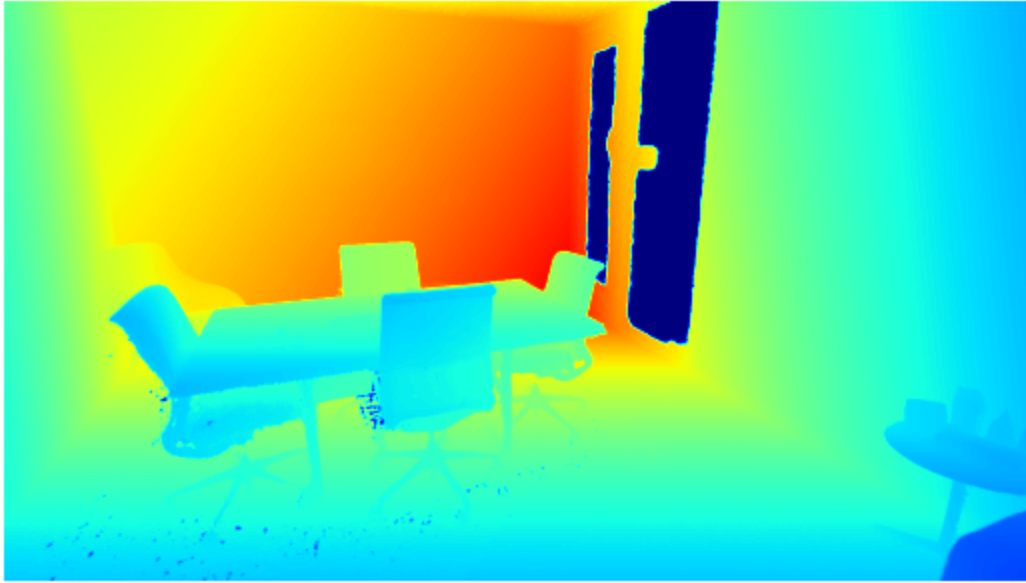}
        \caption{Rendered depth}
        \label{fig:figure4}
    \end{subfigure}
    \caption{Reconstruction results of Office 2 scene from Replica dataset.}
    \label{fig:recon}
\end{figure}

Figure \ref{fig:recon} shows the qualitative results of our 3D reconstruction of the ``Office 2" scene from the Replica dataset \cite{replica19}, comparing ground truth images with our rendered outputs. Using half of the original 2000 frames, our RGB and depth renderings closely match the ground truths, accurately capturing the scene's colors, object placement, and spatial depth, highlighting the precision of our reconstruction.

\begin{figure}[ht!]
    \centering
    \begin{subfigure}[b]{0.49\columnwidth}
        \centering
        \includegraphics[width=\textwidth]{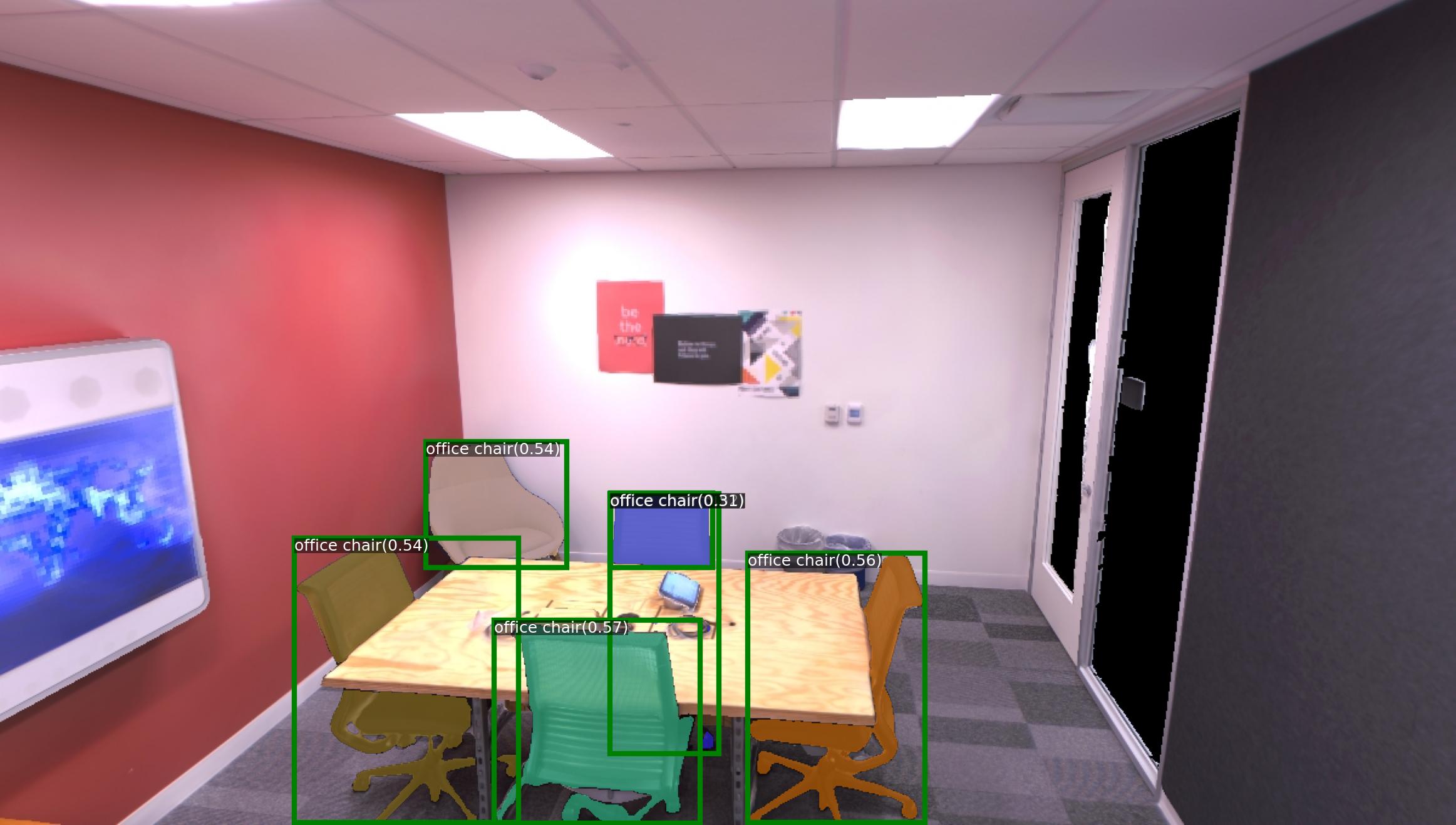}
        \caption{Segmented ``office chair"}
        \label{fig:figure1}
    \end{subfigure}
    \hfill
    \begin{subfigure}[b]{0.49\columnwidth}
        \centering
        \includegraphics[width=\textwidth]{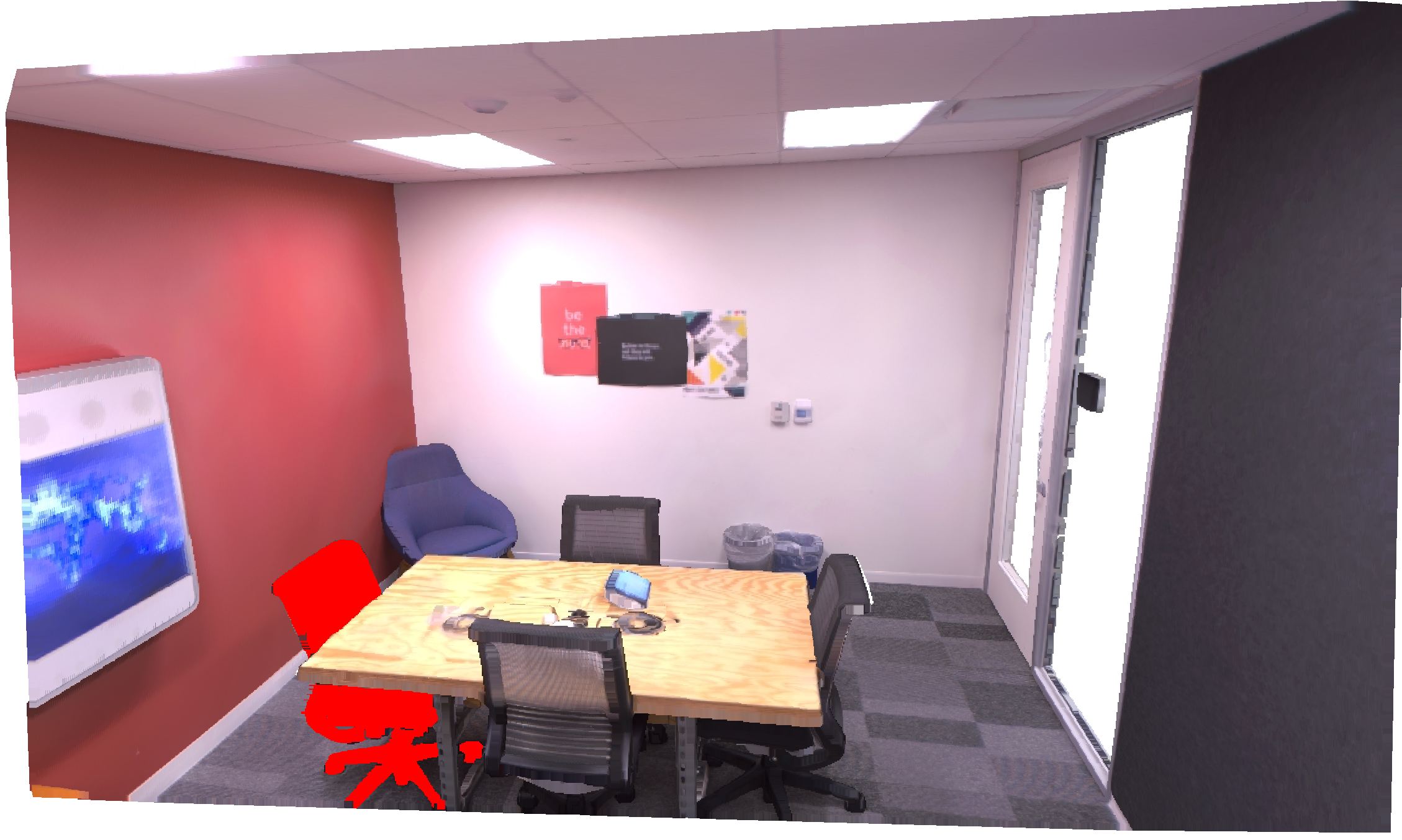}
        \caption{Localized point cloud}
        \label{fig:figure2}
    \end{subfigure}
    \vskip\baselineskip
    \begin{subfigure}[b]{0.49\columnwidth}
        \centering
        \includegraphics[width=\textwidth]{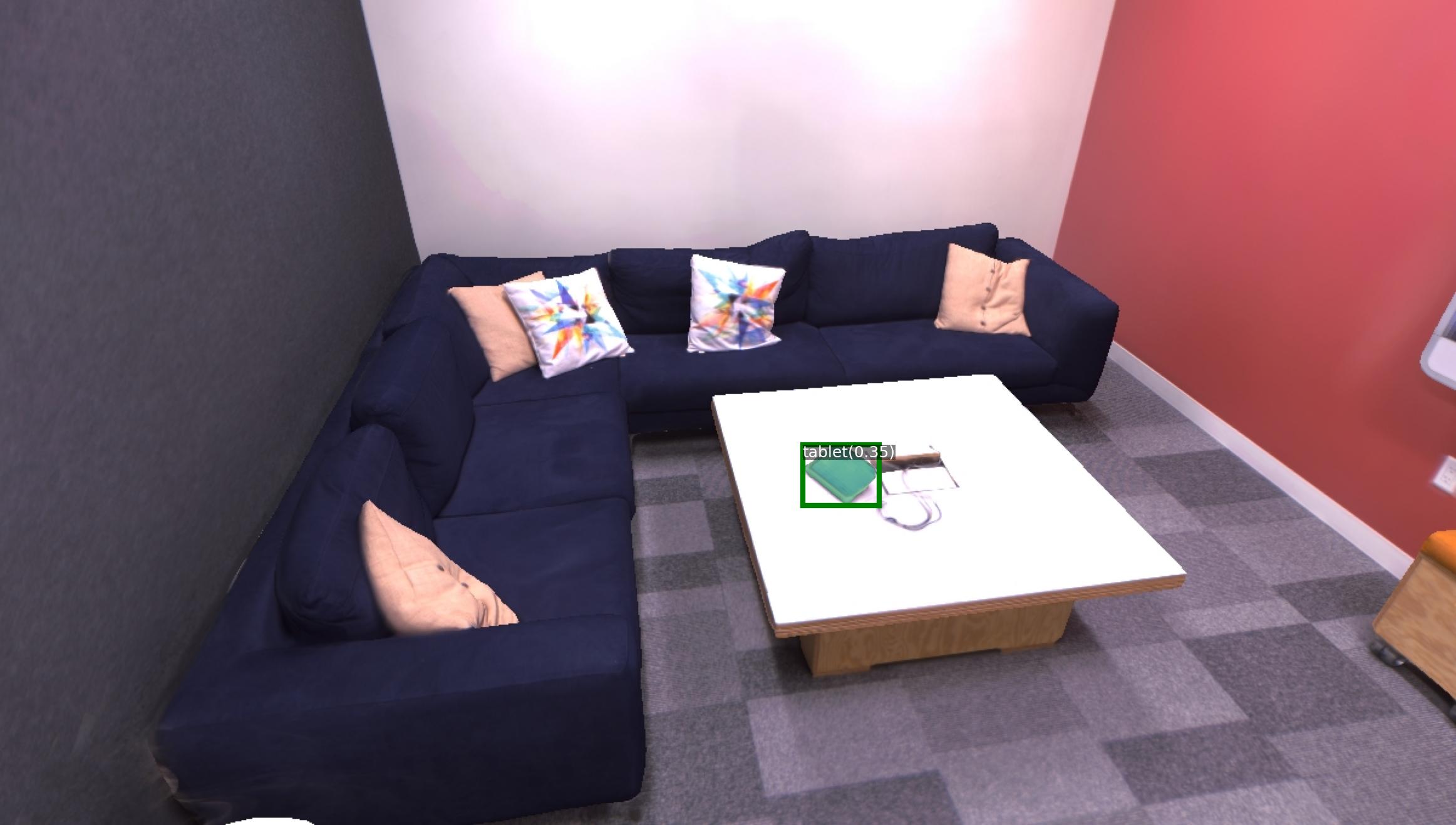}
        \caption{Segmented ``tablet"}
        \label{fig:figure3}
    \end{subfigure}
    \hfill
    \begin{subfigure}[b]{0.49\columnwidth}
        \centering
        \includegraphics[width=\textwidth]{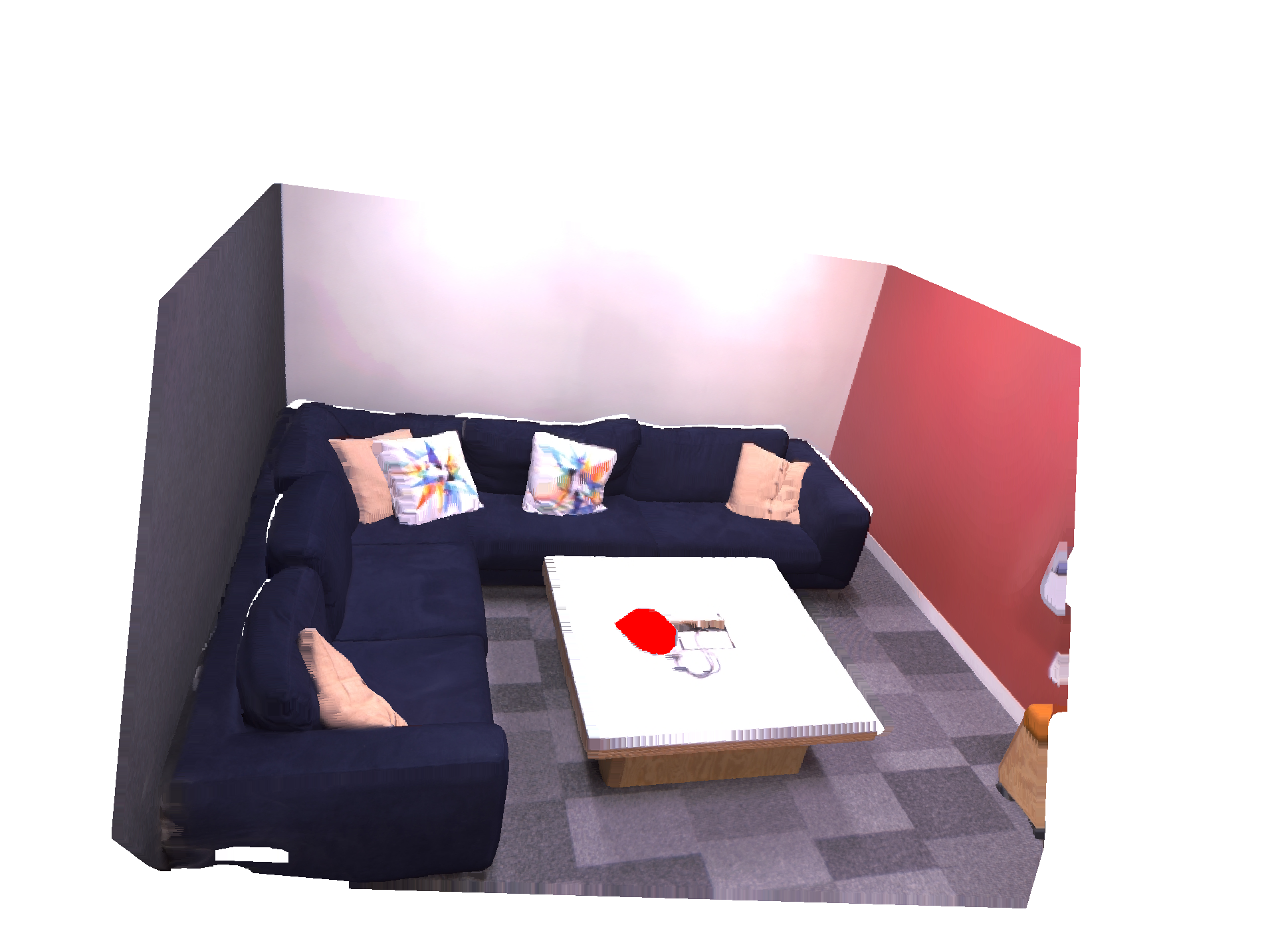}
        \caption{Localized point cloud}
        \label{fig:figure4}
    \end{subfigure}
    \caption{Visualization of the localized object with the queries ``Office chair" and ``Tablet"}
    \label{fig:pcd}
\end{figure}

Figure \ref{fig:pcd} illustrates the 3D reconstruction of an ``Office 2" scene, highlighting the localization in response to the ``Office chair" and ``Tablet" queries. The queried objects, emphasized in red, are effectively identified from other elements within the scene, showcasing the capability of the system to identify queried objects accurately.

\subsubsection{Quantitative results}

Figure \ref{tab:quantitative} displays the precision, recall, and Intersection over Union (IoU) metrics comparing our Go-SLAM method with two baseline approaches across three different environments: Office 2, Room 0, and Room 2. For each environment, the models were tasked with localizing 10 distinct objects within the scene.

In Office 2, Go-SLAM achieved a precision of 0.61, a recall of 0.73, and an IoU of 0.59, surpassing both Baseline 1 (0.38 precision, 0.42 recall, 0.33 IoU) and Baseline 2 (0.54 precision, 0.58 recall, 0.44 IoU). Similar trends are observed in Room 0 and Room 2, where Go-SLAM consistently outperforms the baselines in all metrics. The improvements over Baseline 1 demonstrate the robustness of our grounded object segmentation, particularly with challenging objects in natural settings. Additionally, the improvements over Baseline 2 highlight the effectiveness of our object localization techniques.


\section{CONCLUSION}

In conclusion, Go-SLAM introduces a novel approach to 3D scene reconstruction, combining Gaussian Splatting SLAM with state-of-the-art object segmentation and open-vocabulary querying. Our framework successfully integrates 3D reconstruction, object detection, and natural language understanding to enable real-time environmental interactions. Through comprehensive experiments, we demonstrated that Go-SLAM achieves higher precision, recall, and IoU compared to the baseline methods, particularly in handling complex, unknown environments. The system’s ability to seamlessly embed object-level information into the 3D scene allows for flexible object localization and querying. Overall, Go-SLAM represents a significant step forward in SLAM technology, bridging the gap between scene reconstruction and semantic object understanding.

\section*{ACKNOWLEDGMENT}
This material is based upon work supported in part by the DEVCOM Army Research Laboratory under cooperative agreement W911NF2020221.









{\small
\bibliographystyle{IEEEtran}
\bibliography{refs}
}

\end{document}